\pdfoutput=1

\documentclass[11pt]{article}

\usepackage{acl}

\usepackage{times}
\usepackage{latexsym}
\usepackage{csquotes}
\usepackage{amsmath}
\usepackage{graphicx}
\usepackage{subcaption}
\usepackage{silence}
\usepackage{mathtools}
\usepackage{pgfplots,pgfplotstable}
\pgfplotsset{compat=1.17}
\tikzset{
  font={\fontsize{8pt}{10}\selectfont}}

\usepackage{multirow,ctable}

\usepackage[T1]{fontenc}

\usepackage[utf8]{inputenc}

\usepackage{microtype}

\newcommand{\mn}{\textcolor{black}}
\newcommand{\sara}{\textcolor{black}}
\newcommand{\mg}{\textcolor{black}}

\definecolor{darkyellow}{RGB}{251,188,4}
\definecolor{darkgreen}{RGB}{52,168,83}
\definecolor{lightblue}{RGB}{66,133,244}
\definecolor{acqua}{RGB}{70,189,198}

\title{Attention as a Guide for Simultaneous 
Speech Translation}

\author{Sara Papi\textsuperscript{$\Diamond$$\Box$}, Matteo Negri\textsuperscript{$\Diamond$}, Marco Turchi\textsuperscript{$\triangle$} \\
  \textsuperscript{$\Diamond$}Fondazione Bruno Kessler \\
  \textsuperscript{$\Box$}University of Trento \\
  \textsuperscript{$\triangle$}Independent Researcher \\
  \texttt{\{spapi,negri\}@fbk.eu, marco.turchi@gmail.com}}

\begin{document}
\maketitle
\begin{abstract}

In simultaneous speech translation (SimulST), effective policies 
that determine when to write partial translations
are crucial to reach high output quality with low latency. Towards this objective, we propose \textsc{EDAtt} (\textbf{E}ncoder-\textbf{D}ecoder \textbf{Att}ention), an adaptive policy that exploits the attention patterns between audio source and target textual translation to guide an offline-trained ST model during simultaneous inference. \textsc{EDAtt} exploits the attention scores 
modeling the audio-translation relation to decide whether to emit a partial hypothesis or wait for more audio input. This is done under the assumption that, if attention is focused towards the most recently received speech segments, 
the information they provide
can be insufficient to generate the hypothesis (indicating that the system has to wait for additional audio input). Results on en$\rightarrow$\{de, es\} show that \textsc{EDAtt} yields better results compared to the SimulST state of the art, with gains respectively up to 7 and 4 BLEU points for the two languages, and with a reduction in computational-aware latency  up to 1.4$s$  and 0.7$s$  compared to existing SimulST policies applied to offline-trained models.

\end{abstract}

\section{Introduction}

In simultaneous speech translation (SimulST), systems have to generate  translations incrementally while concurrently receiving audio input. 
This requirement poses a significant challenge
since the need of generating high-quality outputs has to be balanced with the need to minimize their latency, i.e. the time elapsed
(lagging)
between when a word is uttered and when it is actually translated by the system. 

In direct SimulST systems \citep{berard_2016,weiss2017sequence},\footnote{\sara{In this paper, we focus on direct models that exhibit lower latency and better performance 
compared to
traditional cascade architectures 
composed
of separate automatic speech recognition and machine translation  components \citep{ansari-etal-2020-findings,iwslt_2021,anastasopoulos-etal-2022-findings}.}}
the balance between output quality and latency
is managed by a 
\textit{decision policy}, which is the strategy for determining, at each time step, whether to emit a partial translation or to wait for additional audio input.
Decision
policies can be divided into two categories: \textit{fixed} and \textit{adaptive}. Fixed policies are usually based on simple heuristics \citep{ma-etal-2019-stacl}, while 
\mg{adaptive policies take into account the actual input content}
to make the decisions \citep{zheng-etal-2020-simultaneous}.
\sara{Recent works \citep{liu-etal-2021-cross,zaidi2021decision,zaidi22_interspeech,zhang2022information} proved the superiority of adaptive policies over fixed ones.
However, \sara{a major limitation}
of these policies is that they} require training \textit{ad-hoc} \sara{and complex} SimulST architectures,
which results in high computational costs.

\mn{Computational costs are also inflated by the common}
practice of simulating 
\sara{the simultaneous test conditions by providing partial input during training}
\sara{to avoid \mn{the} quality 
\mn{drops}}
caused by the mismatch between 
training and
\sara{test}
conditions \citep{ren-etal-2020-simulspeech,ma-etal-2020-simulmt,ma2021streaming,han-etal-2020-end,zeng-etal-2021-realtrans,liu-etal-2021-ustc,zaidi2021decision,zaidi22_interspeech}. This practice is independent of the decision policy 
\mn{adopted, and}
typically requires dedicated trainings for each latency regime. To 
\sara{mitigate}
this issue, offline-trained ST systems have been employed for 
\mn{simultaneous}
inference 
\mn{\citep{liu20s_interspeech,chen-etal-2021-direct,nguyen2021empirical} and, along this direction,} \citet{papi2022does} demonstrated that dedicated trainings simulating the inference conditions are not 
\mn{necessary since}
offline-trained systems outperform those specifically trained for SimulST.
The effectiveness of using offline-trained ST models for simultaneous inference
has been \sara{also} confirmed by
the last IWSLT 2022 evaluation campaign \citep{anastasopoulos-etal-2022-findings}, where 
the winning submission to the SimulST task \citep{polak-etal-2022-cuni}
\sara{is an offline model exploiting}
the Local Agreement policy by \citet{liu20s_interspeech}.
However, despite its good results, 
this policy relies on a strategy (the generation of two consecutive hypotheses prior to starting the emission) that has a significant impact on  latency.
This raises the need for effective  policies that \textit{i)} are adaptive,  \textit{ii)} are directly applicable to offline ST models, and \textit{iii)} achieve low latency at low computational costs.

Towards these objectives,
\mn{we} propose
\textsc{EDAtt} (\textbf{E}ncoder-\textbf{D}ecoder \textbf{Att}ention),\footnote{Code, outputs and offline ST models used for our experiments are released under Apache License 2.0 at: \url{https://github.com/hlt-mt/fbk-fairseq}.} a novel \mn{adaptive} policy for SimulST that leverages the encoder-decoder attention patterns of an offline-trained ST model to decide when to emit partial translations.
In a nutshell, our idea is that 
the next word
of the partial hypothesis
at a given time step is safely emitted only if the system 
does not attend to the most recent audio frames, 
meaning that the information received up to that time step is sufficient to 
generate
that word.
Building on this idea, our contributions are summarized as follows:




\begin{itemize}
\item
We introduce \textsc{EDAtt}, a novel adaptive decision policy for SimulST, which guides offline-trained ST models during simultaneous inference by looking at the
attention patterns 
dynamically 
computed from
the audio input over time;

\item
We show that \textsc{EDAtt} outperforms the Local Agreement policy applied to the same offline ST models at almost all latency regimes, with computational-aware average lagging (AL\_CA) reductions up to 1.4$s$ for German and 0.7$s$ for Spanish on MuST-C \citep{CATTONI2021101155};

\item
We show that \textsc{EDAtt} also outperforms the state-of-the-art
CAAT architecture 
\citep{liu-etal-2021-cross},
especially in terms of AL\_CA, with gains of up to 7.0 BLEU for German and 4.0 BLEU for Spanish.

\end{itemize}

\section{Background}
In terms of architectural choices, Transformer \citep{transformer} and its derivatives \citep{gulati20_interspeech,Chang2020EndtoEndAW,papi-etal-2021-speechformer,efficient-conformer,kim2022squeezeformer,andrusenko2022uconv} 
are the \textit{de-facto} standard both in offline and 
simultaneous ST \citep{ansari-etal-2020-findings,iwslt_2021,anastasopoulos-etal-2022-findings}. 

A generic Transformer model is composed of an encoder, whose role is to map the input speech sequence $\mathbf{X}=[x_1,...,x_n]$ into an internal representation,
and a decoder, whose role is to generate the output textual sequence $\mathbf{Y}=[y_1,...,y_m]$ by exploiting 
the internal representation
in an auto-regressive manner \citep{graves2013generating}, that is by consuming the previously generated output as additional input when generating the next one.

The encoder and the decoder are composed of a stack of identical blocks, whose components may vary depending on the particular Transformer-based architecture,
although they all share the same  
dot-product attention mechanism \citep{7472621}.
In general, the attention is a function that maps a query 
matrix $Q$ and a 
pair
of key-value 
matrices
($K$, $V$) to an output 
matrix
\citep{7472618}. The output is obtained as a weighted sum of $V$, whose weights are computed through a compatibility function between $Q$ and $K$ that, in the case of the scaled dot-product attention used in the original Transformer formulation, is:
\begin{equation*}
    A(Q,K,V) = softmax \left( \frac{QK^T}{\sqrt{d_k}} \right) V
\end{equation*}
where $d_k$ is the dimension of $K$. 
The attention $A$ is 
\mn{computed on}
$h$ heads in parallel, each 
\mn{applying}
learned linear projections $W^Q$, $W^K$, and $W^V$ to the $Q$, $K$, and $V$ matrices. These representations 
are then concatenated and projected using another learned matrix $W^O$, resulting in the final output:
\begin{equation*}
    \begin{multlined}
    \text{Multihead}(Q,K,V) =  \\ \text{Concat}(\text{head}_1, \text{head}_2 ,..., \text{head}_h) W_O 
    \end{multlined}
\end{equation*}
where $\text{head}_i = A(QW^Q_i,KW^K_i,VW^V_i)$.

In the encoder layers, $Q$, $K$, and $V$ are computed from the same speech input sequence $\mathbf{X}$, realizing 
the so-called \textit{self}-attention $A_{\text{self}}(\mathbf{X})$. Differently, in the decoder layer, two types of attention are computed 
sequentially:
self-attention, and 
\textit{encoder-decoder} 
(or cross)
attention.
In the encoder-decoder attention, $Q$ comes from the previous decoder layer (or directly from the previously generated output $\textbf{Y}$, in the case of the first decoder layer) while $K$ and $V$ come from the output of the encoder, hence the matrix can be expressed as $A_\text{cross}(\mathbf{X},\mathbf{Y})$. 
In this 
work,
we 
only exploit the
encoder-decoder attention matrix to guide the model during 
simultaneous
inference. 
Therefore, we use the notation $A$ instead of $A_{\text{cross}}$ for simplicity, and \mn{henceforth} refer to this matrix as the encoder-decoder representation of a 
\mn{specific}
decoder layer $d$ considering the attention head $h$.

\section{\textsc{EDAtt} policy}
\label{sec:policy}

We propose to exploit the information contained in the encoder-decoder attention matrix of an offline ST model during inference
to determine whether to
wait for 
additional
audio input
or 
emit
a partial 
translation.
The
use of attention as the core mechanism of our policy is motivated by related works in machine translation (MT) and language modeling, which prove that attention scores can encode syntactic dependencies \citep{raganato-tiedemann-2018-analysis,htut2019attention} and language representations \citep{Lamarre2022}, as well as align source and target tokens \citep{tang-etal-2018-analysis,zenkel2019adding,garg-etal-2019-jointly,chen-etal-2020-accurate}.
We posit (and demonstrate in Section \ref{sec:attn}) that this encoder-decoder attention relationship between source audio and target tokens also exists in offline ST models, and can be used to guide them during simultaneous inference.

\begin{figure}[tb]
\renewcommand*\thesubfigure{\arabic{subfigure}} 
     \begin{subfigure}[b]{0.48\textwidth}
        \centering
         \includegraphics[width=0.475\textwidth]{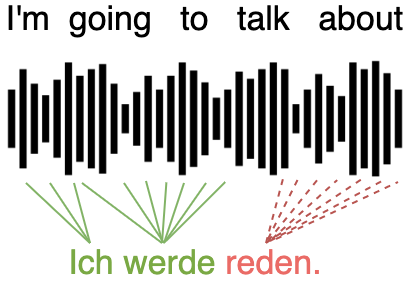}
         \caption{When the first speech segment is received, the partial hypothesis \enquote{\textit{Ich werde}} is emitted since the attention is not concentrated towards the end of the segment while \enquote{\textit{reden.}} is not since the attention is all concentrated on the last frames.}
     \end{subfigure}
     \par\medskip
     \begin{subfigure}[b]{0.48\textwidth}
        \centering
         \includegraphics[width=0.725\textwidth]{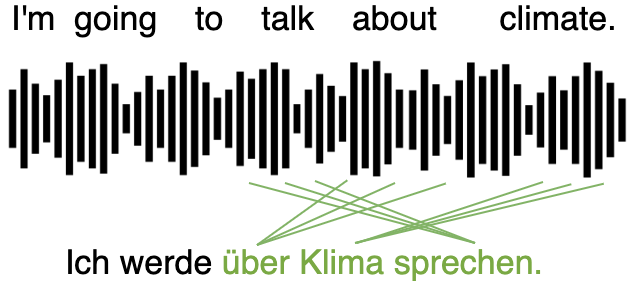}
         \caption{When the second speech segment is received, the new partial hypothesis \enquote{\textit{über Klima sprechen.}} is emitted since the attention is not concentrated towards the end of the segment.}
     \end{subfigure}
    \caption{Example of the \textsc{EDAtt} policy. Links 
    indicate
    where the attention weights point to.}
    \label{fig:edatt}
\end{figure}

\mn{Our approach builds on the following hypothesis (see Figure \ref{fig:edatt}): at each time step, if the attention is focused towards the end of the input audio sequence (1), the system will probably need more information to correctly produce the current output candidate. On the contrary (2), if the attention concentrates on early audio frames (far enough from the last received ones), the current output candidate can be safely emitted because the early encoded information is sufficient. 
Accordingly, the model will continue to emit the next token of the partial hypothesis
until the above condition is 
verified, that is until its encoder-decoder attention scores do not focus towards the end of the received speech segment.
The rationale is that if the encoder-decoder attention of the predicted token points to the most recent speech information -- i.e. attention scores are higher towards the last audio frames received -- this information could
be incomplete and therefore
still insufficient to generate that token.} 


More formally, at each time step $t$, \textsc{EDAtt} determines whether to emit the next token $y_j$,
given the previously generated tokens $\mathbf{Y}_{j-1} = [y_1, ..., y_{j-1}]$ and the partial audio input sequence $\mathbf{X}_t$,
by looking at the sum of the last $\lambda$ encoder-decoder attention weights of the vector $A_j(\mathbf{X}_t, \mathbf{Y}_{j-1})$.
Specifically, $y_j$ is emitted if:
\begin{equation}
\label{eq:attn-threshold}
    \sum_{i=t-\lambda}^{t} A_{i,j}(\mathbf{X}_t, \mathbf{Y}_{j-1}) < \alpha, \quad \alpha \in (0, 1)
\end{equation}
where 
$\alpha$
is a hyperparameter that controls the quality-latency trade-off: lower values of $\alpha$ increase 
\mn{the latency, as they reduce the possibility to satisfy Equation \ref{eq:attn-threshold} (i.e. the sum  of the last $\lambda$ encoder-decoder attention weights will likely exceed $\alpha$), and vice versa.}
When Equation \ref{eq:attn-threshold} is satisfied, $y_j$ is emitted and the same process is repeated
for $y_{j+1}$, and so on. The process continues until we reach the token $y_{j+w}$ for which Equation \ref{eq:attn-threshold} is no longer verified. At that point, the emission is stopped and the total number of tokens emitted at time step $t$ is $w$. 



\section{Experimental Settings}

\subsection{Data}
\label{subsec:data}
To be comparable with previous works \citep{ren-etal-2020-simulspeech,ma-etal-2020-simulmt,zeng-etal-2021-realtrans,liu-etal-2021-cross,papi2022does,zhang2022information}, we train our models on MuST-C en$\rightarrow$\{de, es\} \citep{CATTONI2021101155}.
The choice of the two target languages is also motivated by their different word ordering:
Subject-Object-Verb (SOV) for German and 
Subject-Verb-Object (SVO)
for Spanish.
This opens the possibility of validating our approach on target-language word orderings that are respectively 
different and similar 
with respect to the English (i.e. SVO) source audio.
We also 
perform data augmentation by applying
sequence-level knowledge distillation \citep{kim2016sequencelevel,gaido-2020-on-knowledge,Gaido2022DirectST} as in \citep{liu-etal-2021-cross,papi2022does},
for which the transcripts of MuST-C en$\rightarrow$\{de, es\} are translated with an MT model (more details can be found in Appendix \ref{sec:train_setup}) and used together with the gold reference during training. Data statistics are 
given
in Appendix \ref{sec:data_stat}.

\subsection{Architecture and Training Setup}
\label{sec:architecture}

For our experiments, we use the bug-free implementation by \citet{Papi2023ReproducibilityIN} of the Conformer-based encoder-decoder model for ST \citep{9414858}.
The offline model is made of 12 Conformer encoder layers \citep{gulati20_interspeech} and 6 Transformer decoder layers \sara{($d_{max}=6$)} with a total of $\sim$115M parameters. Each encoder/decoder layer has 8 attention heads \sara{($h_{max}=8$)}. 
The input is represented as 80 audio features extracted every 10$ms$ with sample window of 25 and processed by two 1D convolutional layers with stride 2 to reduce its length by a factor of 4 \citep{wang2020fairseqs2t}. Utterance-level Cepstral Mean and Variance Normalization (CMVN) and SpecAugment \citep{Park2019} are applied during training. 
Detailed 
settings are described in Appendix \ref{sec:train_setup}.

\subsection{Inference and Evaluation}

We use the SimulEval tool \citep{ma-etal-2020-simuleval} to simulate simultaneous conditions and evaluate all the models. 
For our policy, we vary $\alpha$ of Equation~\ref{eq:attn-threshold} in the range $[0.6,0.4,0.2,0.1,0.05,0.03]$ and set the size of the speech segment to 800$ms$.
During inference, the features are computed on the fly and CMVN normalization is based on the global mean and variance estimated on the MuST-C training set. 
All inferences are performed on a single NVIDIA 
K80 GPU with 12GB memory as in the IWSLT Simultaneous evaluation campaigns.

We use sacreBLEU \citep{post-2018-call}\footnote{BLEU+case.mixed+smooth.exp+tok.13a+version.1.5.1} to evaluate translation quality, and Average Lagging \citep{ma-etal-2019-stacl} -- or AL -- to evaluate latency, as in the default SimulEval evaluation setup. As suggested by \citet{ma-etal-2020-simulmt},  
%
%
%
%
for our comparisons with other approaches we also 
report
computational-aware average lagging
%
%
%
%
(AL\_CA), which
measures the real elapsed time instead of the ideal one considered by AL, thus giving a more realistic latency measure when the system operates in real time. 
Its computation is also provided by SimulEval.

\subsection{Terms of Comparison}
\label{subsec:comparison}

We conduct experimental comparisons with 
the 
state-of-the-art 
architecture for SimulST (CAAT) and, respectively, the current best (Local Agreement) and the most widely used (Wait-k) policies that can be directly applied to our offline ST systems for simultaneous inference. In detail:

\paragraph{Cross Attention Augmented Transformer (CAAT) --} 
the 
state-of-the-art 
architecture for SimulST \citep{liu-etal-2021-cross},
winner of the IWSLT 2021 SimulST task \citep{iwslt_2021}.
Inspired by the Recurrent Neural Network Transducer \citep{graves2012sequence}, it is made of three Transformer stacks: the encoder, the predictor, and the joiner.
These three elements are jointly trained to optimize translation quality while keeping latency under control. We train and evaluate the CAAT model using the code provided by the authors,\footnote{\url{https://github.com/danliu2/caat}} and on the same data used for our offline ST model.
 
\paragraph{Local Agreement (LA) --} 
the 
state-of-the-art 
decision policy introduced by \citet{liu20s_interspeech}, and used by the winning system at  IWSLT 2022 \citep{anastasopoulos-etal-2022-findings}.
It  consists in  generating  a partial hypothesis from scratch 
each time a new speech segment is added, and emitting it -- or part of it -- 
if it
coincides with one of those generated in the previous $l$ time steps, where $l$ is a hyperparameter. 
Since \citet{liu20s_interspeech} empirically found that considering only the most recent previously generated tokens ($l=1$) as memory works better, we adopt the same strategy to apply this policy.

\paragraph{Wait-k --}
the 
simplest and 
most 
widely used
decision policy in 
SimulST \citep{ren-etal-2020-simulspeech,ma-etal-2020-simulmt,zeng-etal-2021-realtrans}.
It consists in waiting 
for a fixed number of words
($k$) before starting to emit the translation, and then proceeding by alternating waiting and writing operations. 
Since in SimulST the information about the number of words is not explicitly contained in the audio input, a word detection strategy is used to determine this information. Detection strategies can be fixed, when it is assumed that each word has a pre-defined fixed duration, or adaptive, when the information about the number of words is 
inferred
from the audio content.
Following 
\citet{papi2022does}, we adopt a CTC-based adaptive word detection strategy to detect the number of words.
%
%
%
In addition, to be comparable with the other approaches, we employ beam search to generate each token. 

\begin{figure*}[t]
     \centering
     \begin{subfigure}[b]
     {0.495\textwidth}
         \centering
         \includegraphics[width=\textwidth]{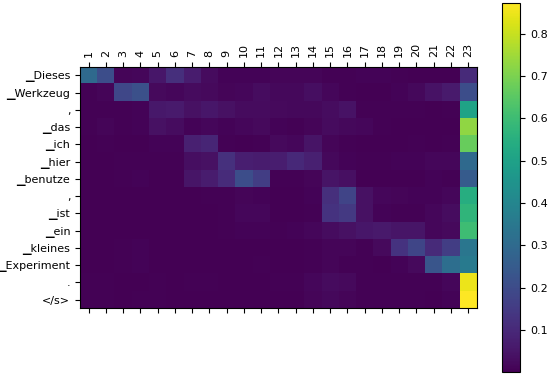}
        \caption{Unfiltered}
     \end{subfigure}
     \hfill
     \begin{subfigure}[b]{0.495\textwidth}
         \centering
         \includegraphics[width=\textwidth]{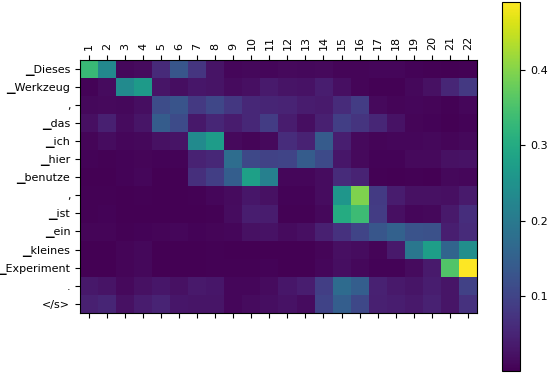}
         \caption{Filtered}
     \end{subfigure}
        \caption{Encoder-decoder attention scores 
        on a random sample of the  
        MuST-C en$\rightarrow$de dev set, before (a) and after (b) the filtering of the last frame from the attention matrix.}
        \label{fig:filtering}
\end{figure*}


\section{Attention Analysis}
\label{sec:attn}

To validate our
hypothesis and study the feasibility of our method, we start by exploring the encoder-decoder attention matrices of the offline trained models.
We proceed as follows: first, by visualizing the attention weights, we check for the existence of patterns that could be exploited during simultaneous inference.
Then, we analyze the performance of the \textsc{EDAtt} policy to discover the best value of $\lambda$, the decoder layer $d$, and the attention head $h$ from which to extract the attention scores that 
better
balance
the quality-latency trade-off. 

\paragraph{Do attention patterns exist also in ST?}
To answer this question, we conducted an analysis of the encoder-decoder matrices obtained from the MuST-C en-de dev
set. Through the visualization of attention weights, we observed a consistent phenomenon across our two language directions (en$\rightarrow$\{de, es\}): the attention weights concentrate on the last frame, regardless of the input length, as shown in Figure 2a.
This 
behaviour
has already been observed in prior works on attention analysis,
showing that attention often concentrates on the initial or final token \citep{clark-etal-2019-bert,kovaleva-etal-2019-revealing,kobayashi-etal-2020-attention,ferrando2022towards}, 
with up to 97\% of attention weights being allocated to these positions. As this might hinder the possibility 
to effectively visualize
attention patterns, similarly to \citep{vig-belinkov-2019-analyzing}, we filtered out the last frame from the attention matrix and then re-normalized it.
In this way, as shown in Figure 2b, we obtained a clear pseudo-diagonal pattern compared to the previous unfiltered representation. 
Such correspondence emerging from the encoder-decoder attention scores after the removal of the last frame  indicates a relationship between the source audio frames and target translation texts 
that can be exploited by our adaptive attention-based policy during simultaneous inference.

\pgfplotstableread[row sep=\\]{
BLEU	AL \\
17.2	1.15 \\
20.4	1.2 \\
22.6	1.5 \\
24.3	1.9 \\
25.0    2.36 \\
}\DEf

\pgfplotstableread[row sep=\\]{
BLEU	AL \\
19.0	1.22 \\
20.5	1.24 \\
23.3	1.57 \\
24.4	1.92 \\
25.1    2.34 \\
}\DEff

\pgfplotstableread[row sep=\\]{
BLEU	AL \\
21.7	1.44 \\
22.9	1.56 \\
24.7	1.88 \\
25.0	2.24 \\
}\DEfff

\pgfplotstableread[row sep=\\]{
BLEU	AL \\
23.4	1.72 \\
24.3	1.85 \\
25.0	2.16 \\
}\DEffff

\pgfplotstableread[row sep=\\]{
BLEU	AL \\
27.5	1.19 \\
29.6	1.38 \\
31.9	1.62 \\
33.5	2.01 \\
34.8    2.5 \\
34.9    2.65 \\
}\ESf

\pgfplotstableread[row sep=\\]{
BLEU	AL \\
30.0	1.38 \\
31.7	1.66 \\
32.8	1.7 \\
34.2	2.06 \\
}\ESff

\pgfplotstableread[row sep=\\]{
BLEU	AL \\
32.8	1.72 \\
33.6	1.79 \\
34.2	2.01 \\
35.2	2.44 \\
}\ESfff

\pgfplotstableread[row sep=\\]{
BLEU	AL \\
33.6	1.92 \\
34.7	2.07 \\
35.2	2.32 \\
}\ESffff

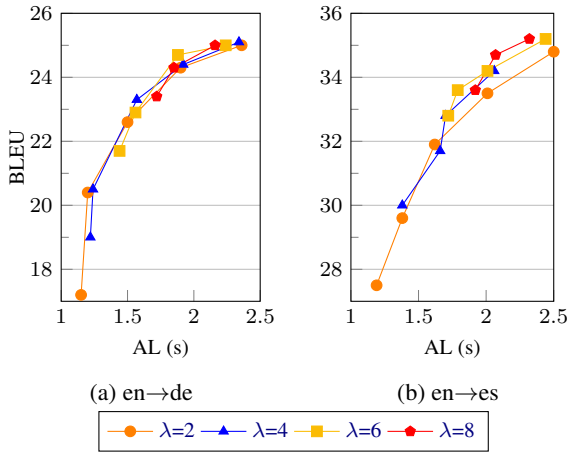
\begin{figure}[!ht]
\centering
\small
\begin{subfigure}[b]{0.23\textwidth}
\begin{tikzpicture}
    \begin{axis}[
            ymajorgrids=true,
            xtick pos=left,
            ytick pos=left,
            minor y tick num=1,
            ytick={18,20,22,24,26},
            ymax=26,
            ymin=17,
            xmax=2.5,
            xmin=1,
            ylabel shift={-4pt},
            ylabel=BLEU, xlabel=AL (s),
            width=4.2cm,
            height=5.4cm,
            xtick=data,
            compat=newest,
            xtick={1,1.5,2,2.5,3},
        ]
        \addplot[color=orange, mark=*] table[x=AL,y=BLEU]{\DEf};
        \addplot[color=blue, mark=triangle*] table[x=AL,y=BLEU]{\DEff};
        \addplot[color=darkyellow, mark=square*] table[x=AL,y=BLEU]{\DEfff};
        \addplot[color=red, mark=pentagon*] table[x=AL,y=BLEU]{\DEffff};
    \end{axis}
\end{tikzpicture}
\caption{en$\rightarrow$de}
\end{subfigure}
\quad
\begin{subfigure}[b]{0.225\textwidth}
\begin{tikzpicture}
    \begin{axis}[
            ymajorgrids=true,
            xtick pos=left,
            ytick pos=left,
            minor y tick num=1,
            ytick={28,30,32,34,36},
            ymax=36,
            ymin=27,
            xmax=2.5,
            xmin=1,
            ylabel=, xlabel=AL (s),
            width=4.25cm,
            height=5.4cm,
            xtick=data,
            compat=newest,
            xtick={1,1.5,2,2.5,3},
            legend style={at={(0.5,-0.2)},    
                    anchor=north,legend columns=4},   
            legend to name={lambda_legend},
        ]
        \addplot[color=orange, mark=*] table[x=AL,y=BLEU]{\ESf};
        \addplot[color=blue, mark=triangle*] table[x=AL,y=BLEU]{\ESff};
        \addplot[color=darkyellow, mark=square*] table[x=AL,y=BLEU]{\ESfff};
        \addplot[color=red, mark=pentagon*] table[x=AL,y=BLEU]{\ESffff};
        \legend{$\lambda$=2, $\lambda$=4, $\lambda$=6, $\lambda$=8}
    \end{axis}
\end{tikzpicture}
\caption{en$\rightarrow$es}
\end{subfigure}
\ref{lambda_legend}
\caption{Effect of $\lambda$ on MuST-C en$\rightarrow$\{de, es\} dev set. We visualize the results with $\text{AL} \leq 2.5s$.}
\label{fig:lambda-num}
\end{figure}

\paragraph{What is the optimal value of $\lambda$?}
To find the best  number of frames ($\lambda$) on which to apply Equation \ref{eq:attn-threshold},
 we analyse the behavior of \textsc{EDAtt} by 
\sara{varying $\alpha$ and setting $\lambda \in [2,4,6,8]$.}\footnote{We do not report the experiments with  $\lambda=1$ since we found that it consistently degrades translation quality. We also experimented with different ways to determine $\lambda$, such as using a percentage instead of a fixed number, but none of them yielded significant differences.}
\sara{For this analysis, we extract the attention scores from the 5\textsuperscript{th} decoder layer ($d=5$) by averaging across the matrices obtained from each attention head ($h=[1,...,8]$) in accordance with the findings of \citep{garg-etal-2019-jointly} about the layer that best represents word alignment.}
We perform the analysis on the MuST-C dev set for both language  
pairs, and present the results in Figure \ref{fig:lambda-num}.
As we can see, as the value of $\lambda$ increases, 
the curves shift towards the right, indicating an increase in latency.
This means that, consistently across languages, considering too many frames towards the end ($\lambda\geq6$) affects latency with little effect on quality.
Since $\lambda=2$  
yields the lowest latency ($\text{AL}\approx1.2s$) in both languages, and especially in
Spanish, we select this value for 
the following experiments.
%
%
\mg{This outcome is noteworthy as it demonstrates that, at least in our settings,
the same optimal value of $\lambda$ applies to diverse target languages with different word ordering.}
However, this might not hold for different source and/or target languages, advocating for future explorations as discussed in the Limitations section.

\pgfplotstableread[row sep=\\]{
BLEU	AL \\
17.4	1.1 \\
17.5	1.18 \\
20.2	1.57 \\
23.0	2.3 \\
}\DEl

\pgfplotstableread[row sep=\\]{
BLEU	AL \\
15.4	1.02 \\
19.1	1.36 \\
23.8	2.18 \\
}\DEll

\pgfplotstableread[row sep=\\]{
BLEU	AL \\
15.8	0.93 \\
17.3	0.98 \\
21.0	1.33 \\
23.1	1.81 \\
24.0    2.35 \\
}\DElll

\pgfplotstableread[row sep=\\]{
BLEU	AL \\
14.7	0.8 \\
15.9	0.83 \\
16.9	0.89 \\
19.3	1.08 \\
22.0    1.41 \\
23.8    1.83 \\
24.6    2.36 \\
}\DEllll

\pgfplotstableread[row sep=\\]{
BLEU	AL \\
17.2	1.15 \\
20.4	1.2 \\
22.6	1.5 \\
24.3	1.9 \\
25.0    2.36 \\
}\DEelllll

\pgfplotstableread[row sep=\\]{
BLEU	AL \\
20.3	1.336 \\
21.0	1.34 \\
23.8	1.74 \\
24.8	2.27 \\
}\DEllllll

\pgfplotstableread[row sep=\\]{
BLEU	AL \\
22.8	1.02 \\
25.9	1.34 \\
27.9	1.56 \\
31.3	2.02 \\
32.5    2.34 \\
}\ESl

\pgfplotstableread[row sep=\\]{
BLEU	AL \\
20.7	0.94 \\
24.9	1.24 \\
27.2	1.49 \\
31.0	2.06 \\
32.5    2.49 \\
}\ESll

\pgfplotstableread[row sep=\\]{
BLEU	AL \\
23.6	1 \\
28.0	1.24 \\
31.1	1.57 \\
34.0	2.14 \\
}\ESlll

\pgfplotstableread[row sep=\\]{
BLEU	AL \\
25.9	1.1 \\
27.6	1.16 \\
29.8	1.32 \\
31.8	1.51 \\
33.6    1.77 \\
34.3    2.07 \\
34.9    2.38 \\
}\ESllll

\pgfplotstableread[row sep=\\]{
BLEU	AL \\
27.5	1.19 \\
29.6	1.31 \\
32.1	1.56 \\
33.7	1.79 \\
34.6    2.06 \\
34.9    2.38 \\
}\ESlllll

\pgfplotstableread[row sep=\\]{
BLEU	AL \\
29.5	1.38 \\
31.9	1.63 \\
33.5	2.01 \\
34.1	2.18 \\
}\ESllllll

\begin{figure}[!t]
\centering
\small
\begin{subfigure}[b]{0.23\textwidth}
\begin{tikzpicture}
    \begin{axis}[
            ymajorgrids=true,
            xtick pos=left,
            ytick pos=left,
            minor y tick num=1,
            ytick={14,16,18,20,22,24,26},
            ymax=26,
            ymin=14,
            xmax=2.5,
            xmin=0.75,
            ylabel shift={-4pt},
            ylabel=BLEU, xlabel=AL (s),
            width=4.2cm,
            height=5.4cm,
            xtick=data,
            compat=newest,
            xtick={0.5,1,1.5,2,2.5},
        ]
        \addplot[color=orange, mark=*] table[x=AL,y=BLEU]{\DEl};
        \addplot[color=lightblue, mark=*] table[x=AL,y=BLEU]{\DEll};
        \addplot[color=darkyellow, mark=*] table[x=AL,y=BLEU]{\DElll};
        \addplot[color=red, mark=*] table[x=AL,y=BLEU]{\DEllll};
        \addplot[color=darkgreen, mark=*] table[x=AL,y=BLEU]{\DEelllll};
        \addplot[color=blue, mark=*] table[x=AL,y=BLEU]{\DEllllll};
    \end{axis}
\end{tikzpicture}
\caption{en$\rightarrow$de}
\end{subfigure}
\quad
\begin{subfigure}[b]{0.225\textwidth}
\begin{tikzpicture}
    \begin{axis}[
            ymajorgrids=true,
            xtick pos=left,
            ytick pos=left,
            minor y tick num=1,
            ytick={20,22,24,26,28,30,32,34,36},
            ymax=36,
            ymin=20,
            xmax=2.5,
            xmin=0.75,
            ylabel=, xlabel=AL (s),
            width=4.3cm,
            height=5.4cm,
            xtick=data,
            compat=newest,
            xtick={0.5,1,1.5,2,2.5},
            legend style={at={(0.5,-0.2)},    
                    anchor=north,legend columns=3},   
            legend to name={dec_lay_legend},
        ]
        \addplot[color=orange, mark=*] table[x=AL,y=BLEU]{\ESl};
        \addplot[color=lightblue, mark=*] table[x=AL,y=BLEU]{\ESll};
        \addplot[color=darkyellow, mark=*] table[x=AL,y=BLEU]{\ESlll};
        \addplot[color=red, mark=*] table[x=AL,y=BLEU]{\ESllll};
        \addplot[color=darkgreen, mark=*] table[x=AL,y=BLEU]{\ESlllll};
        \addplot[color=blue, mark=*] table[x=AL,y=BLEU]{\ESllllll};
        \legend{layer 1, layer 2, layer 3, layer 4, layer 5, layer 6}
    \end{axis}
\end{tikzpicture}
\caption{en$\rightarrow$es}
\end{subfigure}
\ref{dec_lay_legend}
\caption{SimulST results on MuST-C dev set en$\rightarrow$\{de, es\} for each decoder layer $d$. \sara{We visualize the results with $\text{AL} \leq 2.5s$.} 
}
\label{fig:decoder-layer}
\end{figure}
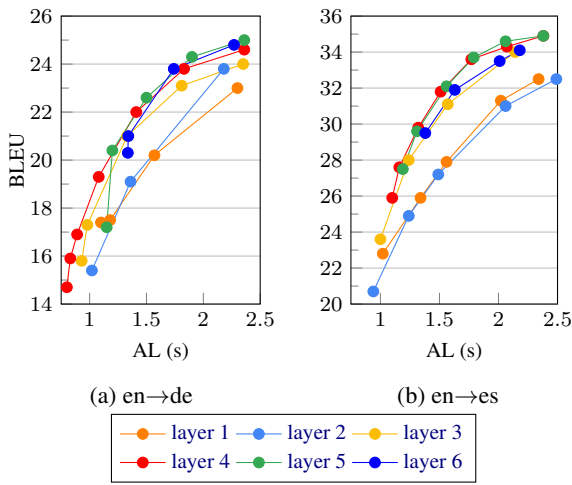

\paragraph{What is the best layer?}
After determining the optimal value of $\lambda$,
we proceed to analyze the \textsc{EDAtt} performance by varying the decoder layer from which the encoder-decoder attention is extracted.
We conduct this study 
by using $\lambda=2$, as 
previously determined
to be the 
optimal
value for both languages.
In Figure \ref{fig:decoder-layer}, we present the 
SimulST results (in terms of AL-BLEU curves)
for each decoder layer \sara{$d=[1,...,6]$}.\footnote{We also tried to make the average of the encoder-decoder attention matrices of each layer but this led to worse results.
}
As we can see, 
on both languages,
Layers 1 and 2 consistently perform worse than the other layers.
Also, Layer 3 achieves inferior quality compared to Layers $\geq$ 4, especially at medium-high latency (AL $\geq$ 1.2$s$)
despite performing better than Layers 1 and 2.
This aligns with the findings of \citet{garg-etal-2019-jointly}, which observed inferior performance by the first three layers \sara{in the alignment task for MT models.}
Concerning Layer 6, both graphs show that the curves cannot achieve lower latency, 
starting
at around 1.5$s$ of AL. This phenomenon is also valid for Layer 5 compared to Layer 4, although being much less pronounced. 
We also observe that Layer 5 
achieves
the best performance at higher latency on both languages.
However, since Layers 5 and 6 never achieve low latency (AL never approaches 1.2$s$), we can conclude that the optimal choice for the simultaneous scenario is Layer 4.  
This is in line
with \citet{Lamarre2022}, which indicates the middle layers as the best choice to provide accurate predictions for language 
representations. 
As a consequence, we will use $d=4$ for the 
subsequent
experiments 
with
\textsc{EDAtt}.

\begin{table}[tb]
    \centering
    \small
    \begin{tabular}{c|ccc|ccc}
        \multirow{2}{*}{\textbf{Head}} & \multicolumn{3}{c|}{\textbf{en$\rightarrow$de}} & \multicolumn{3}{c}{\textbf{en$\rightarrow$es}}\\
        \cline{2-7}
        & 1.2$s$ & 1.6$s$ & 2$s$ & 1.2$s$ & 1.6$s$ & 2$s$ \\
        \hline
         Head 1 & 17.6 & 19.2 & 20.5 & 27.6 & 30.8 & 32.1 \\
         Head 2 & 19.0 & 21.9 & 23.4 & - & 31.9 & 33.9 \\
         Head 3 & - & 22.3 & 23.9 & 27.2 & 29.8 & 31.1 \\
         Head 4 & - & 21.5 & 23.3 & - & 28.4 & 30.7 \\
         Head 5 & 19.2 & 22.2 & 23.8 & - & 30.9 & 32.5 \\
         Head 6 & 18.7 & 21.2 & 22.7 & - & 32.0 & 33.3 \\
         Head 7 & - & 21.9 & 23.5 & - & 30.8 & 32.6 \\
         Head 8 & 19.2 & 20.7 & 21.6 & - & 31.7 & 33.9 \\
         \hline
         Average & \textbf{20.3} & \textbf{22.8} & \textbf{24.0} & \textbf{28.6} & \textbf{32.4} & \textbf{34.1} \\
    \end{tabular}
    \caption{BLEU scores on MuST-C dev set en$\rightarrow$\{de, es\} for each attention head $h$ of Layer 4. Latency (AL) is reported in seconds. \enquote{-} means that the BLEU value is not available or calculable. The last row represents the numerical values of Layer 4 curves of Figure \ref{fig:decoder-layer} obtained by 
    averaging across all 8 heads.}
    \label{tab:attn-heads}
\end{table}


\pgfplotstableread[row sep=\\]{
BLEU	AL \\
19.6	1.43 \\
23.5	2.0 \\
25.1	2.51 \\
25.7	2.97 \\
26.1    3.37 \\
}\ALDEwaitk

\pgfplotstableread[row sep=\\]{
BLEU	AL \\
19.6	2.36 \\
23.5	3.0 \\
25.1	3.53 \\
25.7	4.02 \\
26.1    4.43 \\
}\ALCADEwaitk

\pgfplotstableread[row sep=\\]{
BLEU	AL \\
19.5	1.27 \\
23.1	1.69 \\
24.8	2.04 \\
25.9	2.33 \\
26.4    2.64 \\
}\ALDEla

\pgfplotstableread[row sep=\\]{
BLEU	AL \\
19.5	3.25 \\
23.1	3.32 \\
24.8	3.49 \\
25.9	3.73 \\
26.4    3.98 \\
}\ALCADEla

\pgfplotstableread[row sep=\\]{
BLEU	AL \\
20.3	0.88 \\
20.8	1.32 \\
20.5	1.74 \\
19.9	2.14 \\
19.0    2.54 \\
}\ALDEcaat

\pgfplotstableread[row sep=\\]{
BLEU	AL \\
20.3	1.98 \\
20.8	2.55 \\
20.5	3.14 \\
19.9	3.77 \\
19.0    4.24 \\
}\ALCADEcaat

\pgfplotstableread[row sep=\\]{
BLEU	AL \\
16.8	0.88 \\
19.1	1.04 \\
21.6	1.34 \\
24.0	1.74 \\
25.6    2.26 \\
26.3    2.74 \\
}\ALDEedatt

\pgfplotstableread[row sep=\\]{
BLEU	AL \\
16.8	1.61 \\
19.1	1.75 \\
21.6	2.09 \\
24.0	2.56 \\
25.6    3.26 \\
26.3    3.93 \\
}\ALCADEedatt

\pgfplotstableread[row sep=\\]{
BLEU	AL \\
24.9	1.39 \\
28.4	1.97 \\
29.0	2.5 \\
29.2	2.98 \\
29.4    3.41 \\
}\ALESwaitk

\pgfplotstableread[row sep=\\]{
BLEU	AL \\
24.9	2.41 \\
28.4	3.07 \\
29.0	3.63 \\
29.2	4.09 \\
29.4    4.57 \\
}\ALCAESwaitk

\pgfplotstableread[row sep=\\]{
BLEU	AL \\
22.1	1.12 \\
26.4	1.52 \\
28.1	1.87 \\
28.9	2.17 \\
29.5    2.46 \\
}\ALESla

\pgfplotstableread[row sep=\\]{
BLEU	AL \\
22.1	2.46 \\
26.4	2.56 \\
28.1	2.81 \\
28.9	3.03 \\
29.5    3.28 \\
}\ALCAESla

\pgfplotstableread[row sep=\\]{
BLEU	AL \\
25.1	0.74 \\
26.0	1.15 \\
26.6	1.53 \\
26.6	1.91 \\
26.7    2.27 \\
}\ALEScaat

\pgfplotstableread[row sep=\\]{
BLEU	AL \\
25.1	2.02 \\
26.0	2.57 \\
26.6	3.14 \\
26.6	3.7 \\
26.7    4.25 \\
}\ALCAEScaat

\pgfplotstableread[row sep=\\]{
BLEU	AL \\
23.0	0.95 \\
25.0	1.1 \\
26.6	1.28 \\
27.8	1.52 \\
28.9    1.81 \\
29.2    2.14 \\
}\ALESedatt

\pgfplotstableread[row sep=\\]{
BLEU	AL \\
23.0	1.74 \\
25.0	1.9 \\
26.6	2.09 \\
27.8	2.42 \\
28.9    2.87 \\
29.2    3.37 \\
}\ALCAESedatt

\begin{figure*}[t]
\centering
\small
\begin{subfigure}[b]{0.47\textwidth}
\begin{tikzpicture}
    \begin{axis}[
            ymajorgrids=true,
            xtick pos=left,
            ytick pos=left,
            minor y tick num=1,
            ytick={17,19,21,23,25,27},
            ymax=27,
            ymin=16,
            xmax=5,
            xmin=0.5,
            ylabel=BLEU, xlabel=AL / AL\_CA (s),
            ylabel shift={-3pt},
            width=8.1cm,
            height=6cm,
            xtick=data,
            compat=newest,
            xtick={0.5,1,1.5,2,2.5,3,3.5,4,4.5,5},
        ]
        \addplot[color=orange, mark=pentagon*] table[x=AL,y=BLEU]{\ALDEwaitk};
        \addplot[color=blue, mark=triangle*] table[x=AL,y=BLEU]{\ALDEla};
        \addplot[color=teal, mark=diamond*] table[x=AL,y=BLEU]{\ALDEcaat};
        \addplot[color=red, mark=*] table[x=AL,y=BLEU]{\ALDEedatt};
        \addplot[dashed, color=orange, mark=pentagon*] table[x=AL,y=BLEU]{\ALCADEwaitk};
        \addplot[dashed, color=blue, mark=triangle*] table[x=AL,y=BLEU]{\ALCADEla};
        \addplot[dashed, color=teal, mark=diamond*] table[x=AL,y=BLEU]{\ALCADEcaat};
        \addplot[dashed, color=red, mark=*] table[x=AL,y=BLEU]{\ALCADEedatt};
    \end{axis}
\end{tikzpicture}
\caption{en$\rightarrow$de}
\end{subfigure}
\quad
\begin{subfigure}[b]{0.47\textwidth}
\begin{tikzpicture}
    \begin{axis}[
            ymajorgrids=true,
            xtick pos=left,
            ytick pos=left,
            minor y tick num=1,
            ytick={22,24,26,28,30},
            ymax=30,
            ymin=21,
            xmax=5,
            xmin=0.5,
            ylabel=, xlabel=AL / AL\_CA (s),
            width=8.3cm,
            height=6cm,
            xtick=data,
            compat=newest,
            xtick={0.5,1,1.5,2,2.5,3,3.5,4,4.5,5},
            legend style={at={(0.5,-0.2)},    
                    anchor=north,legend columns=4},   
            legend to name={main_legend},
        ]
        \addplot[color=orange, mark=pentagon*] table[x=AL,y=BLEU]{\ALESwaitk};
        \addplot[color=blue, mark=triangle*] table[x=AL,y=BLEU]{\ALESla};
        \addplot[color=teal, mark=diamond*] table[x=AL,y=BLEU]{\ALEScaat};
        \addplot[color=red, mark=*] table[x=AL,y=BLEU]{\ALESedatt};
        \addplot[dashed, color=orange, mark=pentagon*] table[x=AL,y=BLEU]{\ALCAESwaitk};
        \addplot[dashed, color=blue, mark=triangle*] table[x=AL,y=BLEU]{\ALCAESla};
        \addplot[dashed, color=teal, mark=diamond*] table[x=AL,y=BLEU]{\ALCAEScaat};
        \addplot[dashed, color=red, mark=*] table[x=AL,y=BLEU]{\ALCAESedatt};
        \legend{wait-k, LA, CAAT, EDAtt}
    \end{axis}
\end{tikzpicture}
\caption{en$\rightarrow$es}
\end{subfigure}
\ref{main_legend}
\caption{Comparison with the SimulST systems described in Section \ref{subsec:comparison} on MuST-C en$\rightarrow$\{de, es\} tst-COMMON. Solid curves represent AL, dashed curves represent AL\_CA.}
\label{fig:main-res}
\end{figure*}
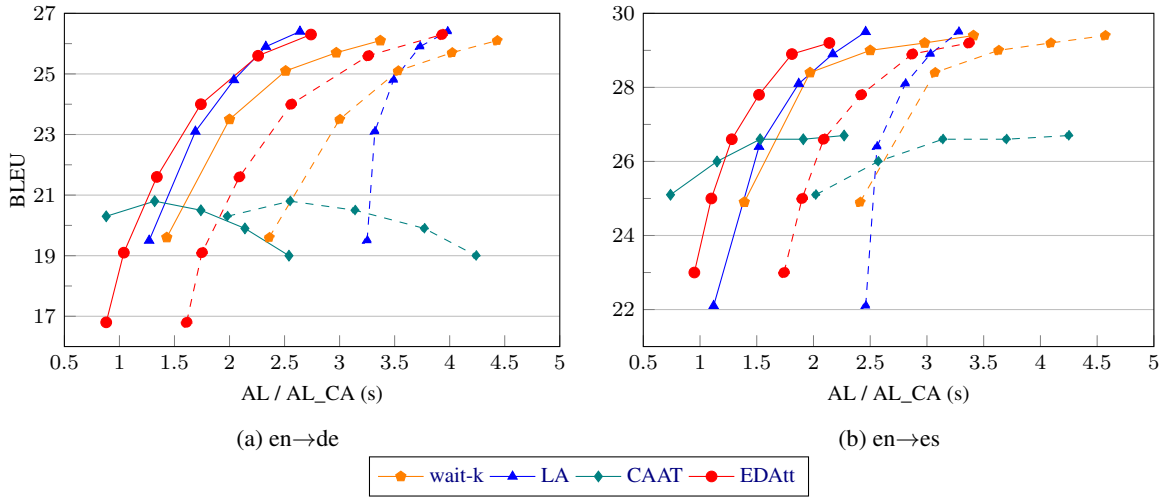

\paragraph{Would a single attention head encode more useful information?}
%
%
According to
prior research 
examining
the usefulness of selecting a single or a set of attention heads to perform natural language processing and translation tasks \citep{jo-myaeng-2020-roles,behnke-heafield-2020-losing,gong2021pay}, we also investigate the behavior of the \textsc{EDAtt} policy by varying the attention head $h$ from which the encoder-decoder attention matrix $A$ is extracted.
In Table \ref{tab:attn-heads},\footnote{A tabular format is used
instead of AL-BLEU curves as many parts of the curves are indistinguishable from each other. AL~=~1.2$s$ is the first latency measure reported because it is the minimum value spanned by the head-wise curves, and AL~=~2$s$ is the last one since increasing latency above this value does not significantly improve translation quality (BLEU).} we present the results obtained from each attention head \sara{$h=[1,...,8]$}.\footnote{Since obtaining a specific latency in
seconds 
is not possible with this method, we 
interpolate the previous and successive points to estimate the BLEU value, when needed.}
Firstly,
we observe that many heads 
are unable to achieve
low latency, 
particularly for Spanish.
%
%
%
Furthermore, there is no consensus on the optimal head among languages or at different latencies (e.g. Head 6 is the best in Spanish at 1.6$s$, but it does not achieve lower latency).
%
%
%
However, we notice that the average across 
all heads (last row) has an overall better performance compared to the encoder-decoder matrices extracted from each individual head, and this holds true for both languages.
%
%
Consequently, we choose to compute the average over the attention heads to apply our \textsc{EDAtt} policy in order to achieve a better quality-latency trade-off for SimulST.


\section{Results}
\label{sec:main_res}

\subsection{Comparison with Other Approaches}

For the comparison of \textsc{EDAtt} with the SimulST systems described in Section \ref{subsec:comparison}, we report in Figure \ref{fig:main-res} both AL (solid curves) and AL\_CA (dashed curves) as latency measures to give a more realistic evaluation of the performance of the systems in real time, as 
recommended
in \citep{ma-etal-2020-simulmt,papi2022does}. %
Results with other metrics, DAL \citep{cherry2019thinking} and LAAL \citep{papi-etal-2022-generation},
are 
provided
in Appendix \ref{sec:DAL_LAAL} for completeness. Numeric values for all the plots are presented in Section \ref{sec:numericvalues}. For our policy, we extract the encoder-decoder attention matrix from Layer 4 ($d=4$), average the weights across heads, and set $\lambda = 2$ 
as it was found to be the optimal setting on the MuST-C dev set for both 
languages,
as 
previously
discussed in Section \ref{sec:attn}.

Quality-latency curves for 
en$\rightarrow$de and en$\rightarrow$es
show similar trends.
The \textsc{EDAtt} policy achieves better overall results compared to the LA and wait-k policies
applied to offline ST models.
\textsc{EDAtt} 
consistently outperforms the wait-k policy, with gains ranging from 1.0 to 2.5 BLEU for German and 1.0 to 3 for Spanish, 
when considering both
ideal (AL) and computationally aware 
(AL\_CA) latency measures.
Additionally, it is able to achieve
lower latency, as the starting point of the wait-k policy is always around 
1.5$s$, while 
\textsc{EDAtt} 
starts at 1.0$s$.
In comparison to the LA policy, we observe an AL\_CA reduction of up to 1.4$s$ for German and 0.7$s$ for Spanish. 
Moreover, the computational overhead of \textsc{EDAtt} is consistently lower, 0.9$s$ on average between languages, against 1.3$s$ of LA. 
Therefore,
the computational cost of our policy is 30\% lower compared to the LA policy. 
Additionally,
\textsc{EDAtt} outperforms 
LA at almost every latency, with gains up to 2.0 BLEU for German and 3.0 for Spanish.

Compared with CAAT, when ideal latency is considered (solid curves), we notice that \textsc{EDAtt} achieves higher quality at medium-high latency (AL $\geq$ 1.2$s$), with BLEU gains up to 5.0 points for German and 2.0 for Spanish. When AL $<$ 1.2$s$, instead, there is a decrease in performance with BLEU drops ranging from 1.5 to 4.0 for German and 1.0 to 2.5 for Spanish.
However, 
when considering
the
realistic 
computational-aware latency measure
AL\_CA (dashed curves), 
we observe 
that the \textsc{EDAtt} curves 
are always to the left of those of the CAAT system,
indicating that our policy always outperforms 
it
with BLEU gains up to 6.0 points for German and 2.0 for Spanish.

In light of this,
we can conclude that
\textsc{EDAtt} achieves new state-of-the-art results 
in terms of
computational-aware metrics, while also being superior at medium-high latency
when considering the
less realistic computational-unaware
measure.

\subsection{Effects of Accelerated Hardware}
\label{subsec:GPU}

To further 
investigate
the computational efficiency of 
\textsc{EDAtt}, 
we 
conducted experiments on all the systems described in
Section \ref{subsec:comparison} using a highly accelerated GPU, an NVIDIA A40 with 48GB memory, during simultaneous inference.

\begin{figure}[t]
\centering
\small
\begin{subfigure}[b]{0.23\textwidth}
\begin{tikzpicture}
    \begin{axis}[
            ymajorgrids=true,
            xtick pos=left,
            ytick pos=left,
            minor y tick num=1,
            ytick={17,19,21,23,25,27},
            ymax=27,
            ymin=16,
            xmax=4.5,
            ylabel shift={-4pt},
            ylabel=BLEU, xlabel=AL\_CA (s),
            width=4.2cm,
            height=5.4cm,
            xtick=data,
            compat=newest,
            xtick={1.5,2,2.5,3,3.5,4,4.5},
        ]
        \addplot[color=orange, mark=pentagon*] table[x=AL,y=BLEU]{\DEwaitk};
        \addplot[color=blue, mark=triangle*] table[x=AL,y=BLEU]{\DEla};
        \addplot[color=teal, mark=diamond*] table[x=AL,y=BLEU]{\DEcaat};
        \addplot[color=red, mark=*] table[x=AL,y=BLEU]{\DEedatt};
    \end{axis}
\end{tikzpicture}
\caption{en$\rightarrow$de}
\end{subfigure}
\quad
\begin{subfigure}[b]{0.225\textwidth}
\begin{tikzpicture}
    \begin{axis}[
            ymajorgrids=true,
            xtick pos=left,
            ytick pos=left,
            minor y tick num=1,
            ytick={20,22,24,26,28,30},
            ymax=30,
            ymin=19,
            xmax=4.5,
            ylabel=, xlabel=AL\_CA (s),
            width=4.2cm,
            height=5.4cm,
            xtick=data,
            compat=newest,
            xtick={1.5,2,2.5,3,3.5,4,4.5},
            legend style={at={(0.5,-0.2)},    
                    anchor=north,legend columns=4},   
            legend to name={main_res_A40_legend},
        ]
        \addplot[color=orange, mark=pentagon*] table[x=AL,y=BLEU]{\ESwaitk};
        \addplot[color=blue, mark=triangle*] table[x=AL,y=BLEU]{\ESla};
        \addplot[color=teal, mark=diamond*] table[x=AL,y=BLEU]{\EScaat};
        \addplot[color=red, mark=*] table[x=AL,y=BLEU]{\ESedatt};
        \legend{wait-k, LA, CAAT, EDAtt}
    \end{axis}
\end{tikzpicture}
\caption{en$\rightarrow$es}
\end{subfigure}
\ref{main_res_A40_legend}
\caption{Effect of using NVIDIA A40 GPU on MuST-C en$\rightarrow$\{de, es\} tst-COMMON considering all the systems of Section \ref{subsec:comparison}. Results are computationally aware.}
\label{fig:main_res_A40}
\end{figure}
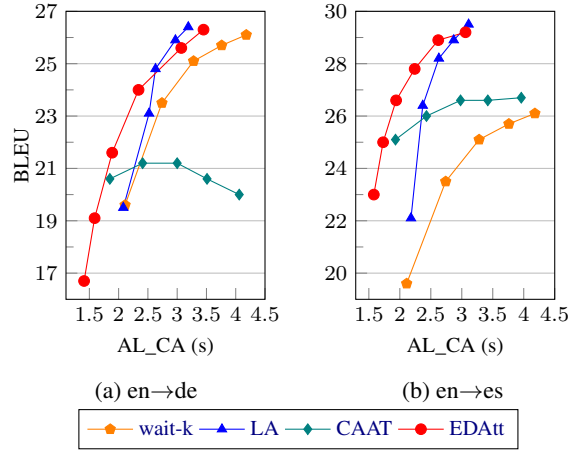

Figure \ref{fig:main_res_A40} reports the results in terms of quality-latency trade-off. 
When comparing the curves
with the computationally aware 
ones in Figure \ref{fig:main-res}
(dashed), it can be observed that
the LA policy seems to benefit more from the use of 
expensive accelerated hardware, 
with
a latency reduction of
0.5-1$s$. However, this reduction is not sufficient to reach a latency lower than 2$s$ 
with this
policy.
Considering the other systems, both wait-k and CAAT curves 
show a slight left shift
(by less than 0.5$s$),
similar to \textsc{EDAtt}.\footnote{
Despite the benefits in terms of quality-latency trade-off, the significantly higher costs of the A40 GPU over the K80 GPU (4.1 vs 0.9 USD/h in Amazon Web Services, \url{https://aws.amazon.com/it/ec2/pricing/on-demand/}) makes 
unlikely
that such a GPU will soon be of widespread use for simultaneous inference.} 

In conclusion, our policy 
proved to be
superior even   
when using accelerated and expensive hardware, further strengthening the 
\sara{previously discussed findings.}
Moreover, these results indicate that there are no significant differences between the systems when using less or more accelerated GPU hardware and advocate for the wider use of computationally aware metrics in future research.



\section{Related Works}

The first policy for SimulST was proposed by \citet{ren-etal-2020-simulspeech} and is 
derived from
the wait-k policy \citep{ma-etal-2019-stacl} 
developed for simultaneous \textit{text-to-text} translation.
%
%
Most of subsequent studies have also adopted the wait-k policy \citep{ma-etal-2020-simulmt,han-etal-2020-end,chen-etal-2021-direct,zeng-etal-2021-realtrans,karakanta-etal-2021-simultaneous,nguyen2021empirical, papi2022does}.
In parallel, several strategies have been developed to directly learn the best policy during training by 
means of \textit{ad-hoc} architectures \citep{ma2021streaming,liu-etal-2021-ustc,liu-etal-2021-cross,chang22f_interspeech} and training procedures aimed at reducing latency \citep{liu-etal-2021-ustc,liu-etal-2021-cross,zaidi2021decision,zaidi22_interspeech,chang22f_interspeech,zhang2022information,omachi2022align}. The latter 
adaptive policies obtained
better performance according to the most recent results observed in \citep{iwslt_2021,anastasopoulos-etal-2022-findings}. We define our policy as adaptive as well, 
as it relies on the encoder-decoder attention mechanism, whose dynamics are influenced by the audio input that increases incrementally over time.
\sara{However, 
\textsc{EDAtt}
completely differs from prior works on adaptive policies that exploit attention
\citep{zaidi2021decision,zaidi22_interspeech,chang22f_interspeech,zhang2022information} because
is the first policy that 
does not require influencing the behaviour of the attention weights through dedicated training strategies, therefore being directly applicable to offline-trained ST models. By doing so, we realize \emph{i)} an adaptive policy, \emph{ii)} directly applicable to offline-trained ST models, \emph{iii)} which achieves low latency at low computational costs.
} 


\pgfplotstableread[row sep=\\]{
BLEU	AL \\
19.6	2.11 \\
23.5	2.74 \\
25.1	3.28 \\
25.7	3.76 \\
26.1    4.18 \\
}\DEwaitk

\pgfplotstableread[row sep=\\]{
BLEU	AL \\
19.5	2.08 \\
23.1	2.52 \\
24.8	2.63 \\
25.9	2.97 \\
26.4    3.19 \\
}\DEla

\pgfplotstableread[row sep=\\]{
BLEU	AL \\
20.6	1.85 \\
21.2	2.41 \\
21.2	3.0 \\
20.6	3.51 \\
20.0    4.06 \\
}\DEcaat

\pgfplotstableread[row sep=\\]{
BLEU	AL \\
16.7	1.41 \\
19.1	1.59 \\
21.6	1.89 \\
24.0	2.34 \\
25.6    3.07 \\
26.3    3.45 \\
}\DEedatt

\pgfplotstableread[row sep=\\]{
BLEU	AL \\
19.6	2.11 \\
23.5	2.74 \\
25.1	3.28 \\
25.7	3.76 \\
26.1    4.18 \\
}\ESwaitk

\pgfplotstableread[row sep=\\]{
BLEU	AL \\
22.1	2.18 \\
26.4	2.37 \\
28.2	2.63 \\
28.9	2.87 \\
29.5    3.11 \\
}\ESla

\pgfplotstableread[row sep=\\]{
BLEU	AL \\
25.1	1.93 \\
26.0	2.43 \\
26.6	2.98 \\
26.6	3.42 \\
26.7    3.96 \\
}\EScaat

\pgfplotstableread[row sep=\\]{
BLEU	AL \\
23.0	1.58 \\
25.0	1.73 \\
26.6	1.94 \\
27.8	2.24 \\
28.9    2.62 \\
29.2    3.06 \\
}\ESedatt

\section{Conclusions}
After investigating
the encoder-decoder attention behavior of offline 
ST
models, we presented \textsc{EDAtt}, a novel adaptive decision policy for 
SimulST that guides an offline ST model to wait or to emit a partial hypothesis by looking at its encoder-decoder attention weights. Comparisons with state-of-the-art SimulST architectures and decision policies reveal that\sara{, at lower computational costs,} \textsc{EDAtt} 
outperforms
the others at almost every latency, with translation quality gains of up to 7.0 BLEU for 
en$\rightarrow$de
and 4.0 BLEU for 
en$\rightarrow$es. 
Moreover, it is also
capable of achieving a computational-aware latency of less than 2$s$ with a reduction of 0.7-1.4$s$ compared to existing decision policies applied to the same offline ST systems. 

\section*{Acknowledgments}
The authors thank Marco Gaido for his valuable support during the paper writing. We acknowledge the support  of the PNRR project FAIR - Future AI Research (PE00000013), under the NRRP MUR program funded by the NextGenerationEU", and of the project \enquote{AI@TN} funded by the Autonomous Province of Trento, Italy.

\section*{Limitations}
Although applicable to any offline ST models, the \textsc{EDAtt} policy and its behavior have been analysed on
models applying CTC compression. Thus, the audio input undergoes a transformation that does not only reduce its dimension but also compresses it into more meaningful units, similar to words or subwords. 
As a consequence, the hyper-parameters regarding the number of frames to which apply the policy ($\lambda$) can vary and depend on the specific ST model. This would require having a validation set on which to search the best value of $\lambda$ before directly testing.
Moreover, the \textsc{EDAtt} policy has been tested on Western European languages and, even if there is no reason suggesting that this cannot be applied (after a proper hyper-parameter search) to other languages, its usage on non-Western European \sara{target} languages \sara{and on a source language different from English} has not been verified in this 
\mn{work and is left for future endeavours.}


\bibliography{custom}
\bibliographystyle{acl_natbib}

\appendix


\pgfplotstableread[row sep=\\]{
BLEU	AL \\
19.6	1.86 \\
23.5	2.42 \\
25.1	2.89 \\
25.7	3.3 \\
26.1    3.66 \\
}\DALDEwaitk

\pgfplotstableread[row sep=\\]{
BLEU	AL \\
19.6	3.14 \\
23.5	3.89 \\
25.1	4.46 \\
25.7	4.95 \\
26.1    5.33 \\
}\DALCADEwaitk

\pgfplotstableread[row sep=\\]{
BLEU	AL \\
19.5	1.98 \\
23.1	2.37 \\
24.8	2.73 \\
25.9	3.01 \\
26.4    3.32 \\
}\DALDEla

\pgfplotstableread[row sep=\\]{
BLEU	AL \\
19.5	7.27 \\
23.1	5.85 \\
24.8	5.37 \\
25.9	5.36 \\
26.4    5.41 \\
}\DALCADEla

\pgfplotstableread[row sep=\\]{
BLEU	AL \\
20.3	1.49 \\
20.8	1.99 \\
20.5	2.46 \\
19.9	2.88 \\
19.0    3.26 \\
}\DALDEcaat

\pgfplotstableread[row sep=\\]{
BLEU	AL \\
20.3	3.28 \\
20.8	3.76 \\
20.5	4.29 \\
19.9	4.86 \\
19.0    5.23 \\
}\DALCADEcaat

\pgfplotstableread[row sep=\\]{
BLEU	AL \\
16.8	1.64 \\
19.1	1.73 \\
21.6	2.01 \\
24.0	2.43 \\
25.6    2.99 \\
26.3    3.46 \\
}\DALDEedatt

\pgfplotstableread[row sep=\\]{
BLEU	AL \\
16.8	2.83 \\
19.1	2.91 \\
21.6	3.26 \\
24.0	3.71 \\
25.6    4.4 \\
26.3    4.97 \\
}\DALCADEedatt

\pgfplotstableread[row sep=\\]{
BLEU	AL \\
24.9	1.96 \\
28.4	2.52 \\
29.0	3.03 \\
29.2	3.45 \\
29.4    3.82 \\
}\DALESwaitk

\pgfplotstableread[row sep=\\]{
BLEU	AL \\
24.9	3.51 \\
28.4	4.3 \\
29.0	4.91 \\
29.2	5.3 \\
29.4    5.73 \\
}\DALCAESwaitk

\pgfplotstableread[row sep=\\]{
BLEU	AL \\
22.1	2.03 \\
26.4	2.42 \\
28.1	2.75 \\
28.9	3.05 \\
29.5    3.33 \\
}\DALESla

\pgfplotstableread[row sep=\\]{
BLEU	AL \\
22.1	4.59 \\
26.4	4.01 \\
28.1	4.1 \\
28.9	4.2 \\
29.5    4.39 \\
}\DALCAESla

\pgfplotstableread[row sep=\\]{
BLEU	AL \\
25.1	1.54 \\
26.0	2.03 \\
26.6	2.51 \\
26.6	2.92 \\
26.7    3.31 \\
}\DALEScaat

\pgfplotstableread[row sep=\\]{
BLEU	AL \\
25.1	3.57 \\
26.0	4.03 \\
26.6	4.54 \\
26.6	5.02 \\
26.7    5.51 \\
}\DALCAEScaat

\pgfplotstableread[row sep=\\]{
BLEU	AL \\
23.0	1.81 \\
25.0	1.92 \\
26.6	2.09 \\
27.8	2.38 \\
28.9    2.74 \\
29.2    3.12 \\
}\DALESedatt

\pgfplotstableread[row sep=\\]{
BLEU	AL \\
23.0	3.01 \\
25.0	3.12 \\
26.6	3.29 \\
27.8	3.62 \\
28.9    4.03 \\
29.2    4.48 \\
}\DALCAESedatt

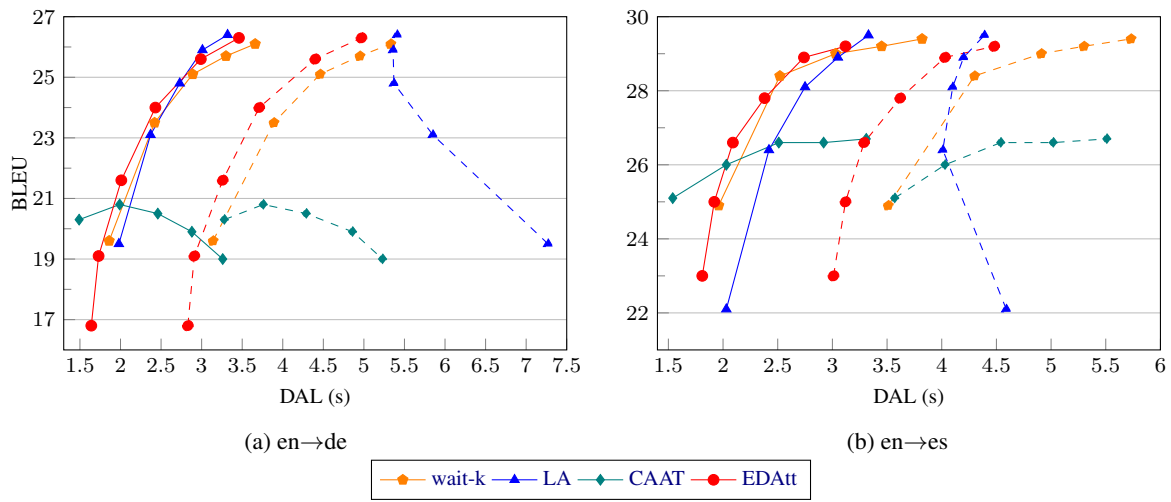
\begin{figure*}[t]
\centering
\small
\begin{subfigure}[b]{0.48\textwidth}
\begin{tikzpicture}
    \begin{axis}[
            ymajorgrids=true,
            xtick pos=left,
            ytick pos=left,
            minor y tick num=1,
            ytick={17,19,21,23,25,27},
            ymax=27,
            ymin=16,
            xmax=7.5,
            xmin=1.3,
            ylabel=BLEU, xlabel=DAL (s),
            ylabel shift={-4pt},
            width=8.2cm,
            height=6cm,
            xtick=data,
            compat=newest,
            xtick={1.5,2,2.5,3,3.5,4,4.5,5,5.5,6,6.5,7,7.5},
        ]
        \addplot[color=orange, mark=pentagon*] table[x=AL,y=BLEU]{\DALDEwaitk};
        \addplot[color=blue, mark=triangle*] table[x=AL,y=BLEU]{\DALDEla};
        \addplot[color=teal, mark=diamond*] table[x=AL,y=BLEU]{\DALDEcaat};
        \addplot[color=red, mark=*] table[x=AL,y=BLEU]{\DALDEedatt};
        \addplot[dashed, color=orange, mark=pentagon*] table[x=AL,y=BLEU]{\DALCADEwaitk};
        \addplot[dashed, color=blue, mark=triangle*] table[x=AL,y=BLEU]{\DALCADEla};
        \addplot[dashed, color=teal, mark=diamond*] table[x=AL,y=BLEU]{\DALCADEcaat};
        \addplot[dashed, color=red, mark=*] table[x=AL,y=BLEU]{\DALCADEedatt};
    \end{axis}
\end{tikzpicture}
\caption{en$\rightarrow$de}
\end{subfigure}
\quad
\begin{subfigure}[b]{0.46\textwidth}
\begin{tikzpicture}
    \begin{axis}[
            ymajorgrids=true,
            xtick pos=left,
            ytick pos=left,
            minor y tick num=1,
            ytick={22,24,26,28,30},
            ymax=30,
            ymin=21,
            xmax=6,
            xmin=1.4,
            ylabel=, xlabel=DAL (s),
            width=8.2cm,
            height=6cm,
            xtick=data,
            compat=newest,
            xtick={1.5,2,2.5,3,3.5,4,4.5,5,5.5,6},
            legend style={at={(0.5,-0.2)},    
                    anchor=north,legend columns=4},   
            legend to name={DAL},
        ]
        \addplot[color=orange, mark=pentagon*] table[x=AL,y=BLEU]{\DALESwaitk};
        \addplot[color=blue, mark=triangle*] table[x=AL,y=BLEU]{\DALESla};
        \addplot[color=teal, mark=diamond*] table[x=AL,y=BLEU]{\DALEScaat};
        \addplot[color=red, mark=*] table[x=AL,y=BLEU]{\DALESedatt};
        \addplot[dashed, color=orange, mark=pentagon*] table[x=AL,y=BLEU]{\DALCAESwaitk};
        \addplot[dashed, color=blue, mark=triangle*] table[x=AL,y=BLEU]{\DALCAESla};
        \addplot[dashed, color=teal, mark=diamond*] table[x=AL,y=BLEU]{\DALCAEScaat};
        \addplot[dashed, color=red, mark=*] table[x=AL,y=BLEU]{\DALCAESedatt};
        \legend{wait-k, LA, CAAT, EDAtt}
    \end{axis}
\end{tikzpicture}
\caption{en$\rightarrow$es}
\end{subfigure}
\ref{DAL}
\caption{DAL results for the SimulST systems of Section \ref{subsec:comparison}. Solid curves represent DAL, dashed curves represent DAL\_CA.}
\label{fig:DAL}
\end{figure*}


\pgfplotstableread[row sep=\\]{
BLEU	AL \\
19.6	1.53 \\
23.5	2.1 \\
25.1	2.6 \\
25.7	3.04 \\
26.1    3.43 \\
}\LAALDEwaitk

\pgfplotstableread[row sep=\\]{
BLEU	AL \\
19.6	2.43 \\
23.5	3.05 \\
25.1	3.57 \\
25.7	4.05 \\
26.1    4.45 \\
}\LAALCADEwaitk

\pgfplotstableread[row sep=\\]{
BLEU	AL \\
19.5	1.41 \\
23.1	1.79 \\
24.8	2.12 \\
25.9	2.39 \\
26.4    2.7 \\
}\LAALDEla

\pgfplotstableread[row sep=\\]{
BLEU	AL \\
19.5	3.31 \\
23.1	3.37 \\
24.8	3.54 \\
25.9	3.77 \\
26.4    4.02 \\
}\LAALCADEla

\pgfplotstableread[row sep=\\]{
BLEU	AL \\
20.3	1.02 \\
20.8	1.40 \\
20.5	1.78 \\
19.9	2.16 \\
19.0    2.54 \\
}\LAALDEcaat

\pgfplotstableread[row sep=\\]{
BLEU	AL \\
20.3	2.09 \\
20.8	2.61 \\
20.5	3.18 \\
19.9	3.78 \\
19.0    4.25 \\
}\LAALCADEcaat

\pgfplotstableread[row sep=\\]{
BLEU	AL \\
16.8	1.08 \\
19.1	1.2 \\
21.6	1.46 \\
24.0	1.83 \\
25.6    2.33 \\
26.3    2.8 \\
}\LAALDEedatt

\pgfplotstableread[row sep=\\]{
BLEU	AL \\
16.8	1.76 \\
19.1	1.87 \\
21.6	2.17 \\
24.0	2.63 \\
25.6    3.31 \\
26.3    3.96 \\
}\LAALCADEedatt

\pgfplotstableread[row sep=\\]{
BLEU	AL \\
24.9	1.58 \\
28.4	2.16 \\
29.0	2.68 \\
29.2	3.14 \\
29.4    3.55 \\
}\LAALESwaitk

\pgfplotstableread[row sep=\\]{
BLEU	AL \\
22.1	1.42 \\
26.4	1.76 \\
28.1	2.08 \\
28.9	2.36 \\
29.5    2.63 \\
}\LAALESla

\pgfplotstableread[row sep=\\]{
BLEU	AL \\
25.1	1.02 \\
26.0	1.37 \\
26.6	1.71 \\
26.6	2.05 \\
26.7    2.38 \\
}\LAALEScaat

\pgfplotstableread[row sep=\\]{
BLEU	AL \\
23.0	1.24 \\
25.0	1.36 \\
26.6	1.52 \\
27.8	1.74 \\
28.9    2.02 \\
29.2    2.34 \\
}\LAALESedatt

\pgfplotstableread[row sep=\\]{
BLEU	AL \\
24.9	2.53 \\
28.4	3.18 \\
29.0	3.72 \\
29.2	4.17 \\
29.4    4.63 \\
}\LAALCAESwaitk

\pgfplotstableread[row sep=\\]{
BLEU	AL \\
22.1	2.65 \\
26.4	2.72 \\
28.1	2.96 \\
28.9	3.17 \\
29.5    3.41 \\
}\LAALCAESla

\pgfplotstableread[row sep=\\]{
BLEU	AL \\
25.1	2.23 \\
26.0	2.72 \\
26.6	3.26 \\
26.6	3.79 \\
26.7    4.33 \\
}\LAALCAEScaat

\pgfplotstableread[row sep=\\]{
BLEU	AL \\
23.0	1.97 \\
25.0	2.1 \\
26.6	2.27 \\
27.8	2.59 \\
28.9    3.01 \\
29.2    3.5 \\
}\LAALCAESedatt

\begin{figure*}[t]
\centering
\small
\begin{subfigure}[b]{0.47\textwidth}
\begin{tikzpicture}
    \begin{axis}[
            ymajorgrids=true,
            xtick pos=left,
            ytick pos=left,
            minor y tick num=1,
            ytick={17,19,21,23,25,27},
            ymax=27,
            ymin=16,
            xmax=4.5,
            xmin=0.9,
            ylabel=BLEU, xlabel=LAAL (s),
            ylabel shift={-4pt},
            width=8.0cm,
            height=6cm,
            xtick=data,
            compat=newest,
            xtick={1,1.5,2,2.5,3,3.5,4,4.5},
        ]
        \addplot[color=orange, mark=pentagon*] table[x=AL,y=BLEU]{\LAALDEwaitk};
        \addplot[color=blue, mark=triangle*] table[x=AL,y=BLEU]{\LAALDEla};
        \addplot[color=teal, mark=diamond*] table[x=AL,y=BLEU]{\LAALDEcaat};
        \addplot[color=red, mark=*] table[x=AL,y=BLEU]{\LAALDEedatt};
        \addplot[dashed, color=orange, mark=pentagon*] table[x=AL,y=BLEU]{\LAALCADEwaitk};
        \addplot[dashed, color=blue, mark=triangle*] table[x=AL,y=BLEU]{\LAALCADEla};
        \addplot[dashed, color=teal, mark=diamond*] table[x=AL,y=BLEU]{\LAALCADEcaat};
        \addplot[dashed, color=red, mark=*] table[x=AL,y=BLEU]{\LAALCADEedatt};
    \end{axis}
\end{tikzpicture}
\caption{en$\rightarrow$de}
\end{subfigure}
\quad
\begin{subfigure}[b]{0.46\textwidth}
\begin{tikzpicture}
    \begin{axis}[
            ymajorgrids=true,
            xtick pos=left,
            ytick pos=left,
            minor y tick num=1,
            ytick={22,24,26,28,30},
            ymax=30,
            ymin=21,
            xmax=4.7,
            xmin=0.9,
            ylabel=, xlabel=LAAL (s),
            width=8.2cm,
            height=6cm,
            xtick=data,
            compat=newest,
            xtick={1,1.5,2,2.5,3,3.5,4,4.5},
            legend style={at={(0.5,-0.2)},    
                    anchor=north,legend columns=4},   
            legend to name={LAAL},
        ]
        \addplot[color=orange, mark=pentagon*] table[x=AL,y=BLEU]{\LAALESwaitk};
        \addplot[color=blue, mark=triangle*] table[x=AL,y=BLEU]{\LAALESla};
        \addplot[color=teal, mark=diamond*] table[x=AL,y=BLEU]{\LAALEScaat};
        \addplot[color=red, mark=*] table[x=AL,y=BLEU]{\LAALESedatt};
        \addplot[dashed, color=orange, mark=pentagon*] table[x=AL,y=BLEU]{\LAALCAESwaitk};
        \addplot[dashed, color=blue, mark=triangle*] table[x=AL,y=BLEU]{\LAALCAESla};
        \addplot[dashed, color=teal, mark=diamond*] table[x=AL,y=BLEU]{\LAALCAEScaat};
        \addplot[dashed, color=red, mark=*] table[x=AL,y=BLEU]{\LAALCAESedatt};
        \legend{wait-k, LA, CAAT, EDAtt}
    \end{axis}
\end{tikzpicture}
\caption{en$\rightarrow$es}
\end{subfigure}
\ref{LAAL}
\caption{LAAL results for the SimulST systems of Section \ref{subsec:comparison}. Solid curves represent LAAL, dashed curves represent LAAL\_CA.}
\label{fig:LAAL}
\end{figure*}
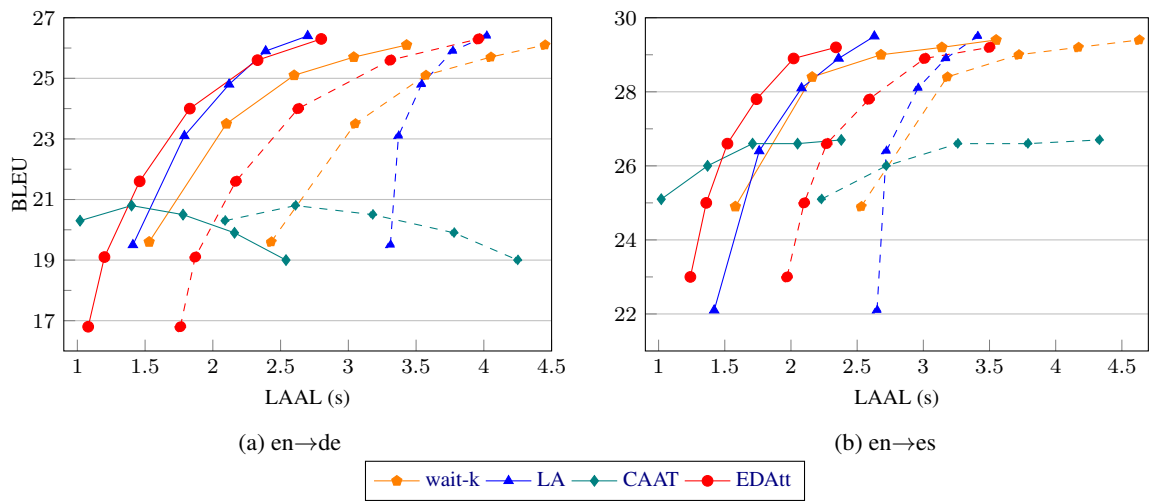
    
\section{Training Settings}
\label{sec:train_setup}

We use 512 as embedding size and 2,048 hidden neurons in the feed-forward layers both in the encoder and in the decoder. 
We set dropout at 0.1 for feed-forward, attention, and convolution layers. Also, in the convolution layer, we set 31 as kernel size for the point- and depth-wise convolutions. 
The vocabularies are based on SentencePiece \citep{sennrich-etal-2016-neural} with dimension of 8,000 \citep{di-gangi-etal-2020-target} for the target side (de, es) and of 5,000 \citep{wang2020fairseqs2t} for the source side (en).
We optimize with Adam \citep{DBLP:journals/corr/KingmaB14} by using the label-smoothed cross-entropy loss with 0.1 as smoothing factor \citep{szegedy2016rethinking}. We employ Connectionist Temporal Classification -- or CTC -- \citep{Graves2006ConnectionistTC} as auxiliary loss to avoid pre-training \citep{gaido-etal-2022-efficient} and also to compress the input audio, reducing RAM consumption and speeding up inference \citep{gaido-etal-2021-ctc}.
The learning rate is set to $5\cdot10^{-3}$ with Noam scheduler \citep{transformer} and warm-up steps of 25k. 
We stop the training after 15 epochs without loss decrease on the dev set and average 7 checkpoints around the best (best, three preceding, and three succeeding). 
Trainings are performed on 4 NVIDIA A40 GPUs with 40GB RAM. We set 40k as the maximum number of tokens per mini-batch, 2 as
update frequency, and 100,000 as maximum updates ($\sim$23 hours).

The MT models used for knowledge distillation are trained on OPUS \citep{tiedemann-2016-opus} en$\rightarrow$\{de, es\} sections and are plain Transformer architectures with 16 attention heads and 1024 embedding features in the encoder/decoder, resulting in $\sim$212M parameters. We achieve 32.1 and 35.8 BLEU on, respectively, MuST-C tst-COMMON German and Spanish.

\section{Data Statistics}
\label{sec:data_stat}

MuST-C training data (train set) 
\sara{has been}
filtered: samples containing audio longer than 30$s$ are discarded to reduce GPU computational requests.
The total number of samples used during our trainings is shown in Table \ref{tab:num_samples}.

\begin{table}[htb]
    \centering
    \begin{tabular}{l|c|cc}
        \textbf{split} & \textbf{en$\rightarrow$de} & \textbf{en$\rightarrow$es} \\
        \hline
         train & 225,277* & 260,049* \\
         dev & 1,423 & 1,316 \\
         tst-COMMON & 1,422 & 1,315\\
    \end{tabular}
    \caption{Number of samples for each split of MuST-C. * means this number doubled due to the use of KD.}
    \label{tab:num_samples}
\end{table}

\section{Main Results with Different Latency Metrics}
\label{sec:DAL_LAAL}
Apart from AL, two metrics can be adopted to measure latency in simultaneous. The first one is the Differentiable Average Lagging -- or DAL -- \citep{cherry2019thinking}, a differentiable version of AL, and the Length-Adaptive Average Lagging -- or LAAL -- \citep{papi-etal-2022-generation}, which is a modified version of AL that accounts also for the case in which the prediction is longer compared to the reference.
Figure \ref{fig:DAL} and \ref{fig:LAAL} show the results of the systems of Figure \ref{fig:main-res} by using, respectively, DAL and LAAL considering both computational aware (CA) and unaware metrics for German and Spanish. 
Numeric values are presented in Section \ref{sec:numericvalues}.

As we can see, the results of Figure \ref{fig:DAL} and \ref{fig:LAAL} confirm the phenomena found in Section \ref{fig:main-res}, indicating \textsc{EDAtt} as the best system among languages and latency values. We observe also that DAL reports higher latency for all systems (it spans from 3 to 7.5$s$ for German and to 5.5$s$ for Spanish), with a counter-intuitive curve for the LA method considering its computational aware version. However, we acknowledge that DAL is less suited than AL/LAAL to evaluate current SimulST systems: 
in its computation, DAL gives a minimum delay for each emitted word while all the systems considered in our analysis can emit more than one word at once, consequently being improperly penalized in the evaluation.

\section{Numeric Values for Main Results}
\label{sec:numericvalues}
Table \ref{tab:numeric_values} on the next page.

\begin{table*}[t]
    \centering
    \begin{tabular}{c|ccccccc}
    \toprule
    \multicolumn{8}{c}{\textbf{en-de}} \\
    \hline
       \hline
       \textbf{Policy} & \textbf{BLEU} & \textbf{AL} & \textbf{AL\_CA }& \textbf{LAAL} & \textbf{LAAL\_CA} & \textbf{DAL} & \textbf{DAL\_CA} \\
       \hline
    \multirow{5}{*}{wait-k} & 19.6 & 1.43 & 2.36 & 1.53 & 2.43 & 1.86 & 3.14 \\
    & 23.5 & 2.00 & 3.00 & 2.10 & 3.05 & 2.42 & 3.89 \\
    & 25.1 & 2.51 & 3.53 & 2.60 & 3.57 & 2.89 & 4.46 \\
    & 25.7 & 2.97 & 4.02 & 3.04 & 4.05 & 3.30 & 4.95 \\
    & 26.1 & 3.37 & 4.43 & 3.43 & 4.45 & 3.66 & 5.33 \\
    \hline
    \multirow{5}{*}{LA} & 19.5 & 1.27 & 3.25 & 1.41 & 3.31 & 1.98 & 7.27 \\
   & 23.1 & 1.69 & 3.32 & 1.79 & 3.37 & 2.37 & 5.85 \\
   & 24.8 & 2.04 & 3.49 & 2.12 & 3.54 & 2.73 & 5.37 \\
   & 25.9 & 2.33 & 3.73 & 2.39 & 3.77 & 3.01 & 5.36 \\
   & 26.4 & 2.64 & 3.98 & 2.70 & 4.02 & 3.32 & 5.41 \\
   \hline
    \multirow{5}{*}{CAAT} & 20.3 & 0.88 & 1.98 & 1.02 & 2.09 & 1.49 & 3.28 \\
    & 20.8 & 1.32 & 2.55 & 1.40 & 2.61 & 1.99 & 3.76 \\
    & 20.5 & 1.74 & 3.14 & 1.78 & 3.18 & 2.46 & 4.29 \\
    & 19.9 & 2.14 & 3.77 & 2.16 & 3.78 & 2.88 & 4.86 \\
    & 19.0 & 2.54 & 4.24 & 2.54 & 4.25 & 3.26 & 5.23 \\
    \hline
   \multirow{6}{*}{\textsc{EDAtt}} & 16.8 & 0.88 & 1.61 & 1.08 & 1.76 & 1.64 & 2.83 \\
   & 19.1 & 1.04 & 1.75 & 1.20 & 1.87 & 1.73 & 2.91 \\
   & 21.6 & 1.34 & 2.09 & 1.46 & 2.17 & 2.01 & 3.26 \\
   & 24.0 & 1.74 & 2.56 & 1.83 & 2.63 & 2.43 & 3.71 \\
   & 25.6 & 2.26 & 3.26 & 2.33 & 3.31 & 2.99 & 4.40 \\
   & 26.3 & 2.74 & 3.93 & 2.80 & 3.96 & 3.46 & 4.97 \\
    \toprule
    \multicolumn{8}{c}{\textbf{en-es}} \\
    \hline
       \hline
       \textbf{Policy} & \textbf{BLEU} & \textbf{AL} & \textbf{AL\_CA }& \textbf{LAAL} & \textbf{LAAL\_CA} & \textbf{DAL} & \textbf{DAL\_CA} \\
       \hline
       \multirow{5}{*}{wait-k} & 24.9 & 1.39 & 2.41 & 1.58 & 2.53 & 1.96 & 3.51 \\
       & 28.4 & 1.97 & 3.07 & 2.16 & 3.18 & 2.52 & 4.30 \\
       & 29.0 & 2.50 & 3.63 & 2.68 & 3.72 & 3.03 & 4.91 \\
       & 29.2 & 2.98 & 4.09 & 3.14 & 4.17 & 3.45 & 5.30 \\
       & 29.4 & 3.41 & 4.57 & 3.55 & 4.63 & 3.82 & 5.73 \\
       \hline
       \multirow{5}{*}{LA} & 22.1 & 1.12 & 2.46 & 1.42 & 2.65 & 2.03 & 4.59 \\
       & 26.4 & 1.52 & 2.56 & 1.76 & 2.72 & 2.42 & 4.01 \\
       & 28.1 & 1.87 & 2.81 & 2.08 & 2.96 & 2.75 & 4.10 \\
       & 28.9 & 2.17 & 3.03 & 2.36 & 3.17 & 3.05 & 4.20 \\
       & 29.5 & 2.46 & 3.28 & 2.63 & 3.41 & 3.33 & 4.39 \\
       \hline
       \multirow{5}{*}{CAAT} & 25.1 & 0.74 & 2.02 & 1.02 & 2.23 & 1.54 & 3.57 \\
       & 26.0 & 1.15 & 2.57 & 1.37 & 2.72 & 2.03 & 4.03 \\
       & 26.6 & 1.53 & 3.14 & 1.71 & 3.26 & 2.51 & 4.54 \\
       & 26.6 & 1.91 & 3.70 & 2.05 & 3.79 & 2.92 & 5.02 \\
       & 26.7 & 2.27 & 4.25 & 2.38 & 4.33 & 3.31 & 5.51 \\
       \hline
       \multirow{6}{*}{\textsc{EDAtt}} & 23.0 & 0.95 & 1.74 & 1.24 & 1.97 & 1.81 & 3.01 \\
       & 25.0 & 1.10 & 1.90 & 1.36 & 2.10 & 1.92 & 3.12 \\
       & 26.6 & 1.28 & 2.09 & 1.52 & 2.27 & 2.09 & 3.29 \\
       & 27.8 & 1.52 & 2.42 & 1.74 & 2.59 & 2.38 & 3.62 \\
       & 28.9 & 1.81 & 2.87 & 2.02 & 3.01 & 2.74 & 4.03 \\
       & 29.2 & 2.14 & 3.37 & 2.34 & 3.50 & 3.12 & 4.48 \\
    \toprule
    \end{tabular}
    \caption{Numeric values for the plots presented in Sections \ref{sec:main_res} and \ref{sec:DAL_LAAL}.}
    \label{tab:numeric_values}
\end{table*}

\end{document}